\pdfoutput=1

\documentclass[11pt]{article}

\usepackage{ACL2023}

\usepackage{times}
\usepackage{latexsym}

\usepackage{amsmath,amsfonts}
\newcommand{\argmax}{\arg\!\min}

\usepackage[T1]{fontenc}

\usepackage[utf8]{inputenc}

\usepackage{microtype}

\usepackage{inconsolata}

\usepackage{microtype}
\usepackage{graphicx}
\usepackage{subfigure}
\usepackage{booktabs} 
\usepackage[linesnumbered,ruled,vlined]{algorithm2e}
\title{Contextual Moral Value Alignment Through Context-Based Aggregation}

\author{Pierre Dognin, Jesus Rios, Ronny Luss, Inkit Padhi, Matthew D Riemer, Miao Liu,\\ {\bf Prasanna Sattigeri, Manish Nagireddy, Kush R. Varshney, Djallel Bouneffouf }\\ 
 \texttt{\bf IBM Thomas J. Watson Research Center}\\}

\begin{document}
\maketitle
\begin{abstract}
Developing value-aligned AI agents is a complex undertaking and an ongoing challenge in the field of AI. Specifically within the domain of Large Language Models (LLMs), the capability to consolidate multiple independently trained dialogue agents, each aligned with a distinct moral value, into a unified system that can adapt to and be aligned with multiple moral values is of paramount importance. In this paper, we propose a system that does contextual moral value alignment based on contextual aggregation. Here, aggregation is defined as the process of integrating a subset of LLM responses that are best suited to respond to a user input, taking into account features extracted from the user's input. 
The proposed system shows better results in term of alignment to human value compared to the state of the art.
\end{abstract}

\section{Introduction}

In an increasingly interconnected world, the alignment of values and intentions among individuals and groups has never been more critical \cite{sun2024trustllm,rodriguez2024towards}. Value alignment refers to the process of ensuring that the goals and behaviors of artificial intelligence (AI) systems are consistent with human values, preferences, and ethical principles \cite{ji2023ai,hendrycks2020aligning}. Achieving value alignment is crucial to mitigate potential risks. This involves designing AI systems that prioritize human values such as fairness, safety and transparency \cite{gabriel_2020,brown21}. 

This paper addresses a problem that we term Contextual Moral-Value Alignment (CMVA) which extends the concept of value alignment by acknowledging the context-dependent nature of ethical considerations in AI systems. CMVA recognizes that ethical principles and values may vary across different contexts and cultures; such value are often ambiguous. 

CMVA allows AI systems to resolve this ambiguity by adapting to the context and offering responses that respect diverse moral viewpoints. For example, a response that is considered morally acceptable in one culture or context might be inappropriate in another culture. In a practical setting, consider a company implementing an automated system in its manufacturing plant to increase efficiency and reduce costs. Decisions made by the system must deal with such value alignment ambiguity because decisions must balance potentially conflicting values: Efficiency versus Employee Well-being. Implementing automation could lead to increased efficiency and cost savings which aligns with the company's goal of maximizing profits. On the other hand, implementing automation could lead to fewer opportunities or employee layoffs, which conflicts with the company's value of supporting employees and ensuring their well-being.

A comprehensive understanding of the context is necessary to provide a better decision regarding whether the company should proceed with implementing automation. Similar context-dependent decisions also apply to chatbots; for example, sales agents must balance the "customer is always right" mantra versus the goal of profiting from the customer. The focus of this paper is on such decisions made by Large Language Models (LLMs). To get this type of capability, we propose a Contextual Moral Value Alignment Generative System (CMVA-GS) that explores how one may harness the power of text aggregation from multiple agents to achieve Contextual Value Alignment. 

CMVA-GS is an approach where models, called Moral Value Agents, are trained independently to address different contexts. These agents contribute answers individually, and corresponding responses, along with a user's morality profile, are aggregated using an aggregator module. This aggregator contextualizes the answers obtained, providing a comprehensive synthesis of moral perspectives.  

\section{Related work}

\citet{safeRLHF_2024} discuss the challenges in balancing performance with the safety of LLMs. Safe Reinforcement Learning from Human Feedback (Safe RLHF) is introduced as a novel algorithm designed to address the tension between helpfulness and harmlessness during LLM training, where the safety concern for LLMs is formalized as a constrained optimization task. \citet{zeng2023diversified} provide a quantitative analysis to verify the existence of diversified preferences in commonly used human feedback datasets. To mitigate the alignment ineffectiveness caused by diversified preferences, a novel Multi-Objective Reward learning method is proposed. The method automatically adjusts the learning gradients across different preference data sources.  \citet{jang2023personalized} study the Reinforcement Learning from Personalized Human Feedback (RLPHF) problem, where LLMs are aligned to multiple preferences by modeling alignment as a Multi-Objective Reinforcement Learning (MORL) problem. Compared to single-objective baselines, they achieve personalized alignment by decomposing preferences into multiple dimensions. To the best of our knowledge, none of the above works consider the problem of contextual value alignment that we formalize below.

\section{Problem Setting}
\label{SingleValueAlignment}

This section formalizes the MVA problem and our extension to the CMVA problem.

\textit{Moral-Value Alignment:} We model the Moral-Value Alignment as a MORL problem which involves finding a policy that optimally balances multiple objectives simultaneously, aligning the agent's actions with a set of desired values.
Let $\mathcal{S}$ denote the state space of a LLM, which includes the current conversation history, the prompt given to the model, and the current sentence being generated.  Let $\mathcal{A}$ denote the action space representing the choices available to the LLM at each step. Actions include selecting which word or token to generate next in the sequence as well as selecting special tokens (e.g., indicating the end of a sentence). Also denote $\mathcal{R} \subseteq \mathbb{R}^N$ as the set of possible rewards, where $N$ is the number of objectives.

The Moral-Value Alignment problem can be modeled as a MORL~\cite{morl_2015}, where the reward vector $r = \langle r_1,  \ldots, r_N \rangle \in \mathcal{R}$ is a Moral-Value vector of rewards functions with each $r_i : \mathcal{S}\times\mathcal{A} \rightarrow \mathcal{R}$ representing the reward for a response (i.e., the action) to a given prompt (i.e., the state) for the $i^{th}$ moral value. Given a policy $\pi: \mathcal{S} \rightarrow \mathcal{P}(\mathcal{A})$, where $\mathcal{P}(\mathcal{A})$ is the set of distributions over actions, the objective in Moral-Value Alignment as a MORL is to find a policy that maximizes a weighted combination of the reward functions, where a desired weights $w$ is known before the optimization.
Formally, we seek a policy $\pi^*$ that maximizes the expected return $ \mathbf{J}_w(\pi)$, where the vector of rewards $r = \langle r_1,  \ldots, r_N \rangle$ 
is projected as $r^T w$. 

\textit{Contextual Moral-Value Alignment (CMVA):} 
In practice, every user has a different desired weight vector. We refer to it as a Moral Profile Vector defined as $c = [c_1, \ldots, c_n]$ where $c_i$ represents the degree to which the individual adheres to the $i^{th}$ moral value.
In CMVA, the objective to be optimized in the MVA problem above is modified to $\mathbf{J}_c(\pi)$ where now the reward to be optimized is given by  $r^T c$. 
Thus, the optimal policy 
$  \pi^*(c) = \arg \max_{\pi} \mathbf{J}_c(\pi) $
depends on the moral profile.

\section{Contextual MVA Generative System}
\label{RewardModels}
\begin{figure}[tb]
\centering
    \includegraphics[width=0.9\linewidth]{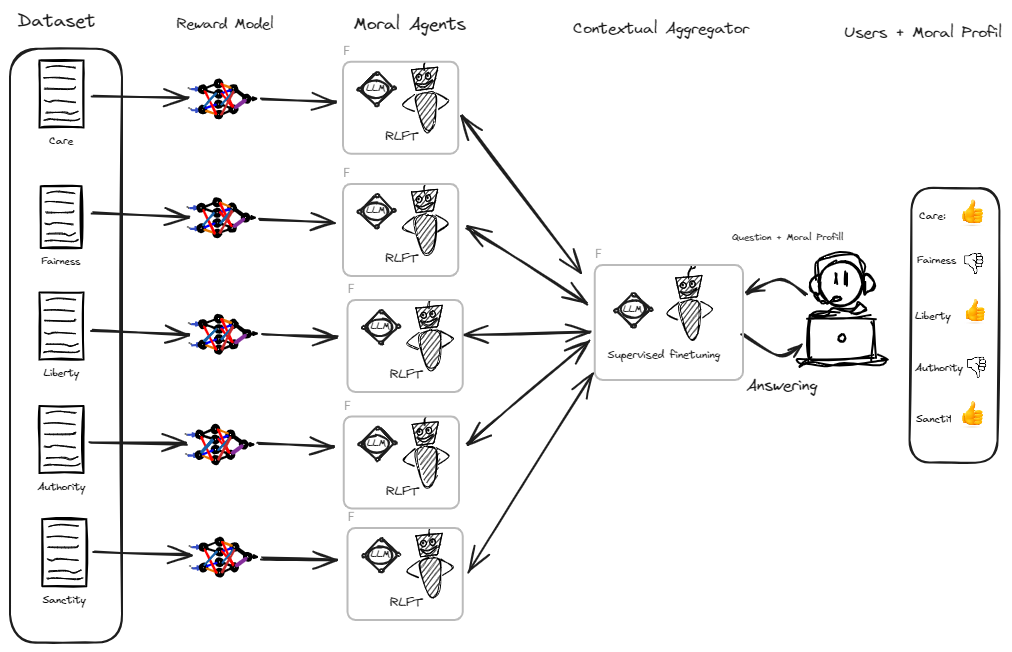}
\caption{Contextual Moral-Value Alignment Generative System
}\label{fig:model}
\end{figure}

Figure~\ref{fig:model} presents the proposed system architecture.
The user request and profile are given as input to the system, through which the request is answered by multiple moral agents. Each agents answers the question according to its moral value. The individualized answers and the user's moral profile are then used by the Contextual Aggregator (CA) to aggregate the answers according to the moral profile. 
The key components of CMVA-GS are:

\textbf{Datasets of Moral Values:} Let $ D = \{(u^{(l)}, z^{(l)})\}_{l=1}^{L} $ be a dataset consisting of $ L $ data points, where: $ u^{(l)} $ represents the features associated with a particular text, $ z^{(l)} $ represents the corresponding moral value  provided by an individual for the $l$-th action within predefined categories representing moral judgments, $ \mathcal{Z} $ denotes the set of possible moral value categories.


\textbf{Reward Models:} We assume $n$ individual values or principles are given, and that we can learn a {\it reward} model $r_i$, for $i=1,\ldots,n$ for each value. A reward is a function that evaluates any LLM's output $y$, i.e., a sequence of tokens generated by the LLM, given a context $x$, with a scalar score representing how much $y$ satisfies the corresponding value or principle. We train one classifier for each moral value (care, fairness, authority, and sanctity) and use these classifiers as reward models to measure how much the output of an LLM aligns with our moral values. Each classifier provides a reward between $0$ and $1$. A reward of $1$ indicates that the output follows the moral value, while a reward of $0$ indicates that it does not follow the moral value.

\textbf{Moral Agents:} At inference, each Moral Agent $\text{A}_i$ takes a question $x$ as input, and outputs an answer $t_i \sim \text{A}_i | x$, for $i=1,\ldots,n$. These answers are then be aggegated according to some moral profile. In order to train each Moral Agent, we use the rewards defined above in Section~\ref{RewardModels} to evaluate the behavior of an LLM which is specified by a policy $\pi$. Specifically, we can measure the LLM's alignment to each moral value $i$ by estimating the expected reward $J_i(\pi)$ by sampling prompts and applying corresponding policies to get responses. 
Optimizing a single-value objective $J_i(\pi)$, i.e., 
\begin{equation} \label{eq:RLFT} 
\pi^*_i = \argmax_{\pi}  J_i(\pi), 
\end{equation} can be done using policy-based Reinforcement Learning (RL) methods such as PPO~\cite{stiennon2020learning,ouyang2022training,schulman2017proximal} with $r_i$ as a reward signal. 

Thus, given a pre-trained LLM to initialize PPO, we can find a LLM $\pi^*_i$ by solving Eq.~\eqref{eq:RLFT} for each value or principle $i=1,\ldots,n$, using RL fine-tuning (RLFT). As a result, we will have $n$ Moral Agents, which are LLMs each denoted by $\text{A}_i$ for $i = 1, \ldots, n$.  To avoid reward hacking during RLFT, a Kullback-Leibler (KL) regularization term can be added to Eq.~\eqref{eq:RLFT} that ensures the policy does not drift too far from its initialization.  

\textbf{Contextual Aggregator:} The CA is a function $\mathcal{M}$ that takes as input, texts and the moral profile feature vector, and produces the output text. The model can be decomposed into an encoder-decoder or decoder-only architecture without loss of generality. Let $\mathcal{E}$ represent the encoder function and $\mathcal{D}$ represent the decoder function. 

Each input text from the moral agents is represented as a sequence of tokens:
$\mathbf{t}_i = (t_{i,1}, t_{i,2}, \dots, t_{i,m_i}) $
where $m_i$ is the length of the $i$-th input text. The moral profile feature vector is denoted as $\mathbf{c}$.
The output text $Y$ is generated by applying the decoder function to the encoded representation of the input texts and moral profile:
$Y = \mathcal{D}(\mathcal{E}(\mathbf{t}_1, \mathbf{t}_2, \dots, \mathbf{t}_n, \mathbf{c}))$.

Let $\mathbf{y} = (y_1, \dots, y_{\ell})$ be the ground truth output text, where $\ell$ is the length of the output text. The loss function $\mathcal{L}$ measures the discrepancy between the generated output $Y$ and the ground truth $\mathbf{y}$. We use the cross-entropy loss:
$\mathcal{L}(\mathbf{y}, Y) = -\sum_{j=1}^{\ell} \sum_{k=1}^{V} y_{j,k} \log(Y_{j,k})$
where $V$ is the size of the vocabulary, $y_{j,k}$ is a one-hot encoding of the $j$-th token in the ground truth output, and $Y_{j,k}$ is the predicted probability of token $k$ at position $j$ in the generated output.

The parameters of the model (i.e., encoder and decoder) are learned by minimizing the loss function using gradient descent-based optimization algorithms:
\begin{align*}
\theta^*= \arg\min_\theta \sum_{i=1}^{N} \mathcal{L}(\mathbf{y}^{(i)}, \mathcal{M}(\mathbf{t}_1^{(i)}, \dots, \mathbf{t}_n^{(i)}, \mathbf{c}^{(i)}))
\end{align*}
where $N$ is the number of training examples, $\theta$ represents the parameters of the model, and $\mathbf{y}^{(i)}$ and $(\mathbf{t}_1^{(i)}, \mathbf{t}_2^{(i)}, \dots, \mathbf{t}_n^{(i)}, \mathbf{c}^{(i)})$ are the ground truth output and input for the $i$-th training example, respectively. For our experimental results, we used a \emph{decoder-only} architecture.
\section{Performance Evaluation}
\label{sec:individual_reward_models}
We present results on the Moral Integrity Corpus (MIC) \cite{ziems-etal-2022-moral}. MIC provides moral annotations on prompt-reply pairs. It was built up from the Social Chemistry (SocialChem) dataset \cite{forbes-etal-2020-social} and shares 5 moral foundations (or values) with it. These 5 values are care-harm, fairness-cheating, loyalty-betrayal, authority-subversion, and sanctity-degradation, defined in Appendix A.4.2 of \citet{forbes-etal-2020-social}. SocialChem annotates each action with a moral judgment that can be binarized capturing negative and neutral/positive judgments to build classifiers ($0$ for negative and $1$ otherwise). These value classifiers can then be used as reward models. Thus, a reward is the probability of a LLM simulated response being in the good class of a moral value classifier.
%

\subsection{Learned Moral Agents} 
\label{sec:learned_moral_agents}

We learn a Moral Agent for each of the $5$ values under consideration. As described in~\ref{SingleValueAlignment}, we start by choosing an initial pre-trained (PT) LLM: Open Assistant $12$B in our case, see PT-model in~\ref{Benchmarks}. Then, we applied RL to fine-tune $5$ times this initial LLM, each time using a different moral reward. We used the PPO implementation from TRL~\cite{vonwerra2022trl}, with a batch size of $256$ episodes (i.e., answers to training questions), $4$ optimization epochs per batch, and a learning rate of $2\!\times\!10^{-9}$.


Table~\ref{tab:individual_models} shows the $5$ learned Moral LLM Agents (one per row). We evaluate their moral behaviour w.r.t. their optimized value by computing the probability (expected reward) that the Moral Agent answers to a dataset of $5$K MIC questions not seen during training follow the individual moral value they are optimized to follow. 
For reference, we provide the probabilities that the PT-model (starting policy in the RLFT of all Moral Agents) follows each individual value. These probabilities are computed using the reward models defined in~\ref{sec:individual_reward_models}.

\begin{table}[h]
    \centering
    \scalebox{0.7}{
    \begin{tabular}{ccc} \toprule
         Moral Value & PT model  & Moral Agents  \\ 
  \midrule
         authority &  91.58\%& 98.83\%\\ 
         fairness      &  85.20\%& 92.40\%\\
         sanctity &  78.37\%& 93.05\%\\
         care                 &  74.70\%& 96.74\%\\
         loyalty         &  74.38\%& 98.20\%\\
 \bottomrule
    \end{tabular}}
    \caption{Probabilities that the Moral Agents answers conform with each moral value.}
    \label{tab:individual_models}
\end{table}

\textit{CMVA-GS: } CMVA-GS models are trained on a dataset derived from MIC. Each input sample is made of a question and moral profile vector from MIC, a context of generated answers from our 5 moral agents.
Models are trained to match the ground-truth answer using cross-entropy. The dataset is composed of $91.0$K/$11.4$K/$11.4$K samples for train/val/test. Our CMVA-GS models start from an OpenAssistant 12B model. We train a Low Rank Adapter (LoRA) \cite{hu2021lora} using supervised fine-tuning. 8 A100-80GB GPUs are used w/ a $5\!\times\!10^{-7}$ learning rate, 128 adapter rank, and 32 per-device minibatch size (256 total). Training runs for $21.6$K steps, a 60 epoch "early stopping".

\subsection{Benchmarks} \label{Benchmarks}

CMVA-GS is compared against the following:

\noindent (1) \textbf{PT-model:} Our PT-model  is the Open-Assistant 12B parameter~\cite{kopf2023openassistant} is a decoder-only model from the Pythia-deduped family~\cite{biderman2023pythia} fine-tuned with (i) Supervised Learning on
QA/dialogue demonstrations as well as (ii) RL on human preferences.

\noindent (2) \textbf{Llama-13b\slash Llama-7b:} We prompt the \textit{Llama-2-13b-chat-hf} and \textit{Llama-2-7b-chat-hf} models, both finetuned to perform dialogue. The prompt defines the desired morals, includes 5 pairs (i.e., questions and answers) of examples per desired moral taken from the MIC (test) data, and requests a response that follows the defined moral values.  

\noindent (3) \textbf{Agg13Llama:} We again prompt the \textit{Llama-2-13b-chat-hf} model but with a twist. The morals are again defined but the examples are the results of passing the user question through the corresponding learned Moral Agents from Section \ref{sec:learned_moral_agents}. The prompt asks the model to aggregate these answers when responding to the question. 

\textit{Evaluation Metrics:} We evaluate our models using ROUGE-1, ROUGE-2, ROUGE-L, and ROUGE-Lsum \cite{lin2004rouge} metrics, widely used in natural language processing to assess the effectiveness of algorithms.

\begin{figure}[tb]
\centering
    \includegraphics[width=0.9\linewidth]{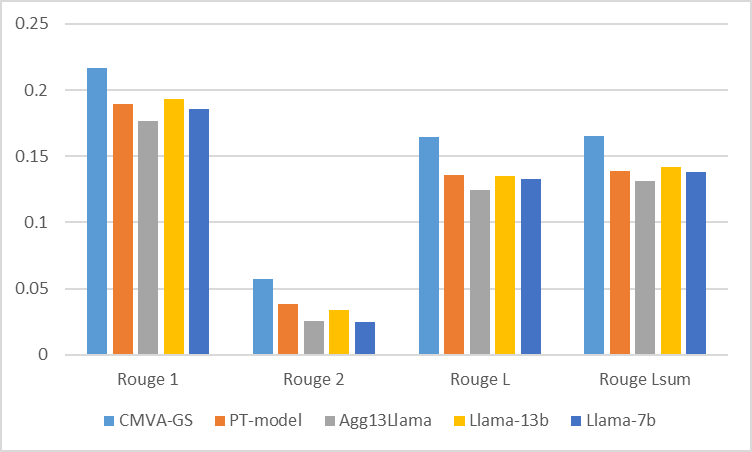}
\caption{Evaluation using ROUGE on MIC. 
}\label{fig:classifier}
\end{figure}

\textit{Experiment Results:} From Figure~\ref{fig:classifier}, we see that
CMVA-GS tends to have the highest ROUGE scores across all metrics, indicating better alignment with human values compared to other models.
PT-model and Llama-13b have similar ROUGE scores, but generally lower than those of CMVA-GS.
Agg13llama has the lowest ROUGE scores among all models, suggesting relatively poorer performance in values alignment.
llama-7b and llama-13b perform better than Agg13llama but still fall short compared to CMVA-GS.
Overall, CMVA-GS appears to be the most effective model in terms of aligning with human values, while the other models vary in their performance, with some showing moderate alignment and others exhibiting relatively lower alignment.

\section{Limitations}
Our paper addresses the contextual value alignment challenge by proposing a novel system that performs contextual aggregation. Our proposed system demonstrates superior results in terms of alignment with human values compared to existing state-of-the-art methods. However, we can highlight multiple limitations of our system:

\textit{Memory:} the system relies on multiple models to perform its task. Running several models concurrently can further exacerbate memory usage and computational overhead. Each additional model adds to the memory footprint and processing requirements of the system.

\textit{User Acceptance and trust:} Users may be hesitant to interact with a system that aggregates responses from multiple sources, especially if they are uncertain about the reliability of the system's behavior.

\textit{Dependency on Training Data Quality:} The effectiveness of the system heavily relies on the quality and representativeness of the training data used for training each individual dialogue agent. If the training data is biased, incomplete, or unrepresentative of diverse perspectives, the aggregated system may inherit these limitations, leading to sub-optimal alignment with human values. If the initial dataset used for training is incorrect, this can create a feedback loop where the RL agent learns from its own mistakes, exacerbating the problem over time.

\bibliography{anthology,custom}
\bibliographystyle{acl_natbib}

\appendix



\end{document}